\documentclass[conference]{IEEEtran}
\IEEEoverridecommandlockouts
\usepackage{cite}
\usepackage{amsmath,amssymb,amsfonts}
\usepackage{graphicx}
\usepackage{textcomp}
\usepackage{xcolor}
\usepackage{subcaption}
\usepackage{pgfplotstable}
\usepackage[linesnumbered,ruled,vlined,noend]{algorithm2e}
\usepackage{algpseudocode}  
\usepackage{amsmath}  
\usepackage{graphicx} 
\usepackage{multirow}
\usepackage[flushleft]{threeparttable}

\usepackage{color}
\usepackage{xspace}
\definecolor{ScarletRed}{rgb}{0.80,0.00,0.00}

\makeatletter
\newcommand{\removelatexerror}{\let\@latex@error\@gobble}
\makeatother

\renewcommand{\arraystretch}{0.1}

\def\BibTeX{{\rm B\kern-.05em{\sc i\kern-.025em b}\kern-.08em
    T\kern-.1667em\lower.7ex\hbox{E}\kern-.125emX}}
    
\begin{document}

\title{Multi-Source Transfer Learning for Non-Stationary Environments\\
}

\author{\IEEEauthorblockN{Honghui Du}
\IEEEauthorblockA{\textit{Department of Informatics} \\
\textit{University of Leicester}\\
Leicester, United Kingdom \\
hd168@leicester.ac.uk}

\and
\IEEEauthorblockN{Leandro L. Minku$^*$}
\IEEEauthorblockA{\textit{School of Computer Science} \\
\textit{University of Birmingham}\\
Birmingham, United Kingdom \\
L.L.Minku@cs.bham.ac.uk
\thanks{$^*$ The corresponding author.}
}

\and
\IEEEauthorblockN{Huiyu Zhou}
\IEEEauthorblockA{\textit{Department of Informatics} \\
\textit{University of Leicester}\\
Leicester, United Kingdom \\
hz143@leicester.ac.uk}

}

\maketitle

\begin{abstract}
In data stream mining, predictive models typically suffer drops in predictive performance due to concept drift. As enough data representing the new concept must be collected for the new concept to be well learnt, the predictive performance of existing models usually takes some time to recover from concept drift. To speed up recovery from concept drift and improve predictive performance in data stream mining, this work proposes a novel approach called Multi-sourcE onLine TrAnsfer learning for Non-statIonary Environments (Melanie). Melanie is the first approach able to transfer knowledge between multiple data streaming sources in non-stationary environments. It creates several sub-classifiers to learn different aspects from different source and target concepts over time. 
The sub-classifiers that match the current target concept well are identified, and used to compose an ensemble for predicting examples from the target concept. We evaluate Melanie on several synthetic data streams containing different types of concept drift and on real world data streams. The results indicate that Melanie can deal with a variety drifts and improve predictive performance over existing data stream learning algorithms by making use of multiple sources.
\end{abstract}

\begin{IEEEkeywords}
concept drift, non-stationary environment, multi-sources, transfer learning.
\end{IEEEkeywords}

\section{Introduction}
Many real world applications produce data in a streaming fashion, i.e., as a sequence of observations that arrive over time. Examples include prediction of customer behaviour, credit card approval, fraud detection, software effort estimation, software defect prediction, etc. A challenge in data stream mining is how to describe a given target probability distribution accurately without knowing the whole data stream beforehand. This challenge is exacerbated by the fact that data streams may suffer from \textit{concept drifts}, i.e., changes in the underlying joint probability distribution of the problem \cite{pocock2010online}. 
We refer to a given joint probability distribution as a \textit{concept}.

One of the reasons why concept drift exacerbates this challenge is that, when a previously unseen joint probability distribution is encountered, existing approaches depend on the arrival of new data to learn an appropriate model of this new distribution. The accuracy of such approaches tends to be poor during the period of time where insufficient data has been received for training. A possible solution to this issue is to use information learned from different sources to speed up the learning of a new target concept, and thereafter improve the accuracy of the estimation. This is called transfer learning \cite{pan2010survey}. For example, when predicting the behaviour of a given target customer, data on other (source) customers may be helpful to improve predictive performance on the target customer. Therefore, transfer learning has the potential to speed up adaptation to concept drift, improving predictive performance in data stream mining.

However, transfer learning is typically defined as an offline learning approach, and almost no work investigates transfer learning in non-stationary data streaming environments \cite{minku2019transfer}. No existing approach can transfer knowledge from different data streaming sources to a given data streaming target in non-stationary environments. And yet, applications where the target domain produces a data stream would typically have source domains that also produce data streams. For example, when predicting customer behaviours, both the source and target customers can produce data streams. This paper thus investigates the following research question: \textit{can multi-source transfer learning improve the predictive performance in data stream mining? When and why?} The assumption is that both the source and target domains produce data streams and may suffer from concept drift.

To answer this question, we propose a novel approach called Multi-sourcE onLine TrAnsfer learning for Non-statIonary Environments (Melanie). Melanie uses online ensemble learning to produce sub-classifiers (base learners) that can represent different parts of the source and target joint probability distributions. When a new joint probability distribution has to be learned (e.g., in the beginning of the learning or after a concept drift), Melanie can transfer knowledge from sub-classifiers that are found to be relevant to the new distribution to improve predictive performance. Experiments show that Melanie can improve predictive performance after concept drifts and can quickly obtain good performance at the early learning stage, when there are few target training examples available.

The paper is organised as follows. Section \ref{related work} introduces related work. Section \ref{Problem Statement} presents the problem statement. Section \ref{Online Multi-Sources Drift Learning} explains the proposed approach Melanie. Section \ref{experiment setup} presents the experimental setup. Section \ref{experiment results} analyses Melanie's predictive performance with synthetic and real world data streams, and compares it with existing approaches. Section \ref{conclusion} presents conclusions and future work. 


\section{Related Work} 
\label{related work}
Sections \ref{Transfer Learning approaches}, \ref{Learning approaches for non-stationary environment} and \ref{Transfer Learning in non-stationary environments} discuss the three main types of approach related to this work.

\subsection{Transfer Learning}
\label{Transfer Learning approaches}

Transfer learning is typically defined for offline learning problems \cite{pan2010survey}. Let $D_S = \{ \left(x_S^{(i)}, y_S^{(i)} \right) \}_{i=1}^N$ denote a data set from a source domain $\mathcal{D}_S = \{ \mathcal{X}_S, p_S(\textbf{x})  \}$ and source task $\mathcal{T}_S = \{\mathcal{Y}_S, p_S(y | \textbf{x}) \}$, where $x_S^{(i)} \in \mathcal{X}_S$, $y_S^{(i)} \in \mathcal{Y}_S$, $\mathcal{X}_S$ is the input space, $\mathcal{Y}_S$ is the output space, $p_S(\textbf{x})$ is the marginal probability distribution and $p_S(y | \textbf{x}) \}$ is the posterior probability distribution. Similarly, define the target data, domain and task as follows: $D_T= \{ (x_T^{(i)}, y_T^{(i)}\}_{i=1}^M$,  $\mathcal{D}_T =  \{ \mathcal{X}, p_T(\textbf{x})  \}$, and $\mathcal{T}_T = \{\mathcal{Y}_T, p_T(y | \textbf{x}) \}$. 
The goal of transfer learning is to use the knowledge learnt from the source to improve the predictive performance of a predictive model $h_T(x): \mathcal{X} \rightarrow \mathcal{Y}$ for the target, despite the fact that the source and target tasks and domains may differ. \textit{Transductive transfer learning} approaches transfer knowledge when $\mathcal{D}_S \neq \mathcal{D}_T$ and $\mathcal{T}_S = \mathcal{T}_T$. \textit{Inductive transfer learning} approaches transfer knowledge between different tasks (e.g. $\mathcal{T}_S \neq \mathcal{T}_T$) while $\mathcal{D}_S = \mathcal{D}_T$ or $\mathcal{D}_S \neq \mathcal{D}_T$. The single source domain definition can be extended to multi-source \cite{minku2019transfer, weiss2016survey}. In some situations, transfer learning may have a negative impact on target learning. This is referred to as negative transfer, and is one of the big challenges in transfer learning \cite{pan2010survey}.

Transfer learning approaches can also be divided into four categories \cite{pan2010survey, weiss2016survey}: instance transfer, feature-representation transfer, relational-knowledge transfer and parameter transfer. 
Among them, parameter transfer approaches share parameters or priors between the source and target. A well known example is TaskTradaBoost \cite{yao2010boosting}. It re-weights sub-classifiers learnt on the source concept by their performance on the target concept. This is particularly interesting in the context of this paper, because it enables knowledge to be transferred through sub-classifiers. This could potentially be used to eliminate the need for storing training examples, which is desirable when dealing with data streams. However, this potential is not exploited by TaskTradaBoost, which still an offline learning approach that requires the whole training set to be available beforehand.


Overall, offline transfer learning approaches  require data sets to be available beforehand, being impractical for dealing with data streams. None of them have automated procedures to continuously learn over time and adapt to concept drifts that may affect the target and source concepts when dealing with data streams.  

\subsection{Data Stream Learning in Non-stationary Environments}
\label{Learning approaches for non-stationary environment}

A data stream is a sequence $D = \{(x^{(t)},y^{(t)})\}_{t=1}^\infty$, where $(x^{(t)},y^{(t)}) \sim p^{(t)}(x,y)$, $x^{(t)} \in \mathcal{X}$ and $y^{(t)} \in \mathcal{Y}$. Data stream learning uses $D$ to train a sequence of predictive models $f^{(t)}(x): \mathcal{X} \rightarrow \mathcal{Y}$ able to generalise to unseen examples from $p^{(t)}(x,y)$. In online learning, at each time $t$, a machine learning algorithm only has access to $f^{t-1}$ and $(x^{(t)},y^{(t)})$ to create $f^t$. This paper concentrates on online learning -- its efficiency makes it more adequate for  applications where multiple data streams need to be processed, as is the case presented in this paper. 



Data streams are often generated by non-stationary environments, which are environments where concept drift may occur \cite{ditzler2015learning, minku2019transfer}. Approaches for data stream learning in non-stationary environments that are able to learn example-by-example rather than chunk-by-chunk \cite{krawczyk2017ensemble} are particularly interesting in the context of this paper, as 
several of them are online learning approaches \cite{minku2012ddd, kolter2007dynamic}. Such approaches can be further divided into active and passive approaches \cite{ditzler2015learning,krawczyk2017ensemble}. Active approaches trigger adaptation mechanisms such as the creation of new models from scratch when concept drift defection methods trigger drift alarms \cite{ditzler2015learning}. 
An example of state-of-the-art active approach is Adaptive Random Forest (ARF) \cite{gomes2017adaptive}.
Examples of drift detection methods include Drift Detection Method (DDM) \cite{gama2004learning}, 
and Early Concept Drift Detection (ECDD) \cite{ross2012exponentially}. 
Passive approaches adopt mechanisms to continuously adapt to any drifts that the environment may suffer, without relying on concept drift detection methods \cite{ditzler2015learning,krawczyk2017ensemble}. 
A popular approach is Dynamic Weighted Majority (DWM) \cite{kolter2007dynamic}.  

Despite having mechanisms to learn data streams, none of these approaches perform transfer learning. In particular, none of them are able to operate in multi-source scenarios.

\subsection{Data Stream Transfer Learning in Non-stationary Environments}
\label{Transfer Learning in non-stationary environments}

Very few approaches perform transfer learning in non-stationary environments \cite{minku2019transfer}. An example is the online inductive parameter transfer learning approach Dynamic Cross-company Mapped Model Learning (Dycom) \cite{minku2014make}. It creates different offline models for different sources, and an online learning model for the target. Each source model is associated to a function that maps predictions made by the source models to the target concept. This function is learnt in an online way and is able to react to concept drift. However, Dycom assumes that only the target arrives in the form of a data stream that may suffer concept drift; the sources are trained offline.


Other two online inductive parameter transfer learning approaches are Diversity for Dealing with Drifts (DDD) \cite{minku2012ddd} and Online Window Adjustment Algorithm (OWA) \cite{zhao2014online}. DDD uses a very highly diverse ensemble to transfer knowledge from the old concept. OWA transfers knowledge from the old concept through a weighted average of the old and new models. However, neither DDD nor OWA can benefit from different sources. Knowledge can only be transferred from the immediate previous target state to the current target state. 

Recently, a new chunk-based inductive parameter transfer approach called Diversity and Transfer-based Ensemble Learning (DTEL) was proposed \cite{sun2018concept}. 
It transfers the structure of a decision tree created with an old chunk of data to the concept represented by the new chunk. It assumes that the old structure is relevant to the new concept. Similar to DDD, it does not consider different sources, with the transfer occurring between a single previous target concept and the new target concept. 
In addition, this is a chunk-based approach, presenting the common chunk-based problems of delaying update to concept drift until a whole new chunk of data received, and assuming that a whole chunk of data always belongs to the same concept.


\section{Problem Statement}
\label{Problem Statement}


This paper tackles the problem of transferring knowledge from one or more sources ($\mathcal{D}_{S_i}, \mathcal{T}_{S_i}$, $i \in \mathbb{N}$) to a target ($\mathcal{D}_T, \mathcal{T}_T$), where the sources and target are represented by data streams, instead of fixed data sets. The data streams come from non-stationary environments, where the distributions underlying $\mathcal{D}_{S_i}$, $\mathcal{D}_T$, $\mathcal{T}_{S_i}$, $\mathcal{T}_T$ may suffer concept drift. Therefore, the aim of the transfer is to improve predictive performance especially during the initial learning stage and after concept drift, when there is little target data to learn from. We will investigate inductive transfer learning, as concept drifts may cause changes in $\mathcal{T}_T$ and $\mathcal{T}_{S_i}$ over time. 

\section{The Proposed Algorithm}
\label{Online Multi-Sources Drift Learning}

In this section, we present our proposed algorithm  Multi-sourcE onLine TrAnsfer learning for Non-statIonary Environments (Melanie). 
%
Melanie is the first approach able to transfer knowledge from both multi-sources and old concepts at the same time, where both sources and target are represented by data streams that may suffer concept drift. It achieves that by using an online inductive parameter transfer strategy. 

Melanie considers that a given source or target concept is composed of several different sub-concepts. We define a source sub-concept as a sub-area of the source input space $\mathcal{X_S'} \in X_S$ associated to its task $\mathcal{T_S'}: \mathcal{X_S'} \rightarrow \mathcal{Y}$. A target sub-concept can be defined in a similar way. Melanie's general idea is to maintain different sub-classifiers (base learners) that may better represent different source and target sub-concepts. When learning a new target joint probability distribution (e.g., in the beginning of the learning or after a concept drift), Melanie identifies which existing sub-classifiers match this new distribution well, i.e., which sub-classifiers represent sub-concepts that share similarities with the new distribution's sub-concepts. These sub-classifiers are then used to transfer knowledge from previously seen source or target distributions to learn the new target distribution more efficiently.



\begin{algorithm}[t]
\caption{Multi-sourcE onLine TrAnsfer learning for Non-statIonary Environments (Melanie)}
\label{pro.alg}
\KwIn{$\left(\textbf{x}^{(t)}, y^{(t)} \right) \in D_i  \quad i \in \{S_1,S_2, \cdots, S_n, T\}$; $M$, set of already seen sources or target, initialised with $\emptyset$; Time forgetting factor $0 \leq \theta \leq 1$; Parameter $0 \leq \delta \leq 1$; Performance index $0 \leq \lambda \leq 1$; Online Learning approach ensemble size $K$; Classifier pool $H_i = \emptyset$ }
\If{$i \notin M$}{ \label{inital.start}
$M \leftarrow M \cup i$ \\
$J_i \leftarrow 1$ (Initialise number of online learning ensembles associated to $i$)\\
Initialise online learning ensemble $H_i^1$\\\label{inital.end}
$H_i \leftarrow H_i \cup H_i^1$ \label{addpool}\\
$\L_{h^{J_i,k}_i} \leftarrow 0$, $\alpha_{h^{J_i,k}_i} \leftarrow 0$, $\forall k < K$ \\\label{lin.initweights}
}

\If{$DriftDetection_i \left(\textbf{x}^{(t)}, y^{(t)} \right) = true$}{ \label{lin.driftdetection}
Initialise a new online learning ensemble $H_{i}^{J_i+1}$\\
$J_i \leftarrow J_i + 1$\\
$\L_{h^{J_i,k}_i} \leftarrow 0$, $\alpha_{h^{J_i,k}_i} \leftarrow 0$, $\forall k < K$ \\\label{lin.initweights2}
$H_i \leftarrow H_i \cup H_{i}^{J_i}$\\ \label{lin.classifiers pool}
\If{$ i = T$}{
$\L_{h^{j,k}_{i'}} \leftarrow 0$, $\alpha_{h^{j,k}_{i'}} \leftarrow 0$, $\forall i' \in M$, $j < J_{i'}$, $k < K$\\\label{lin.drfit.reweight}}

}
OnlineLearningApproach$ \{H_{i}^{J_i},  \left(\textbf{x}^{(t)}, y^{(t)} \right) \} $\\ \label{lin.onlineboosting.target domain}
\If{$ i = T$}{ \label{lin.target received}

\For{all $i' \in M$, $j \leq J_{i'}$, $k \leq K$}{\label{lin.assign weight.target.start}

Calculate the loss of each sub-classifier $L(h^{j,k}_{i'}(\textbf{x}^{(t)})) = \sum_{y^{(j)} \neq y^{(t)}} argmax (0, P(y^{(j)}) - P(y^{(t)}) + \delta ) \quad $ where $P(y)$ is the predicted probability of $h^{j,k}_{i'}(\textbf{x}^{(t)}) = y$  \label{lin.loss of each subclassifiers}

Compute each sub-classifier performance with time forgetting factor
$A_{h^{j,k}_{i'}}^{(t)} = \theta A_{h^{j,k}_{i'}}^{(t - 1)} + (1 - L(h^{j,k}_{i'}(\textbf{x}^{(t)}))) $  \label{lin.time forgetting factor}

$\alpha_{h^{j,k}_{i'}}^{(t)}  = \frac{1}{\sum_{t' = 1}^t \theta^{(t'-1)}}A_{h^{j,k}_{i'}}^{(t)}$ \label{lin.alpha}

}\label{lin.drift.target.end}
\For{all $i' \in M$, $j \leq J_{i'}$, $k \leq K$}{
$\omega_{h^{j,k}_{i'}} =$
\resizebox{1.01\hsize}{!}{$\begin{cases}
\frac{1}{\sum_{{i''}\in\mathcal{M}}\sum_{{j'}=1}^{J_{i'}}\sum_{k=1}^{K} (\alpha_{h^{j',k}_{i''}}^{(t)}>\lambda ? \ \alpha_{h^{j',k}_{i''}}^{(t)}:0)}\alpha_{h^{j,k}_{i'}}^{(t)},& \alpha_{h^{j,k}_{i'}}^{(t)} \geq \lambda \\
0, & otherwise
\end{cases}$}  \label{lin.assign.weight.target.end}
where (testCondition ? v1 : v2) retrieves v1 if testCondition is true, and v2 otherwise.
}
}
\end{algorithm}

Melanie's pseudocode is shown in Algorithm \ref{pro.alg}. 
When an example $\left(x^{(t)}, y^{(t)} \right)$ from a new source or the target $i$ is received for the first time, Melanie creates one online learning ensemble for this source or target (line \ref{inital.start} to \ref{inital.end}). Any online learning ensemble can potentially be used, e.g., online boosting or online bagging \cite{oza2005online}. The idea is that the diversity of the sub-classifiers of such ensembles will cause them to better represent different sub-concepts, facilitating the identification of sub-classifiers whose knowledge could be transferred to the current target.
In the pseudocode, we use the index $i$ to refer to any source or target, i.e., $i \in \{S_1,S_2, \cdots, S_n, T\}$. Therefore, Melanie will have generated $n+1$ online learning ensembles $H_i^1$ in total after all sources and target have generated at least one training example. The set $M$ contains the indexes of all sources and targets for which an online learning ensemble has already been generated.

Each ensemble $H_i^1$ is composed of $K$ sub-classifiers $h^{1,k}_i$, where $1\leq k \leq K$. Line \ref{lin.initweights} is used to initialise the weights associated to each sub-classifier. These weights will be used to identify which sub-classifiers currently match the target distribution well.
Each source and target $i$ is associated to a pool of online learning ensembles $H_i$. The newly created ensemble $H_i^1$ is added to its corresponding pool $H_i$ (line \ref{addpool}). This pool will receive additional ensembles when $i$ suffers concept drift, as explained next. Therefore, each source/target is associated to a pool of ensembles, where each ensemble may represent a different concept observed in that source/target.

Each time a new training example $\left(x^{(t)}, y^{(t)} \right)$ of the source or target $i$ is received, the system  runs a concept drift detection method for $i$ (line \ref{lin.driftdetection}).  Any drift detection method could potentially be used, e.g., DDM \cite{gama2004learning}. If the drift detection method requires monitoring a predictive model representing $i$, the most recent ensemble $H_i^{J_i}$ is used for that.
If a concept drift is detected, Melanie creates a new online learning ensemble $H_i^{J_i+1}$, initialises its weights, and puts it into the pool of ensembles $H_i$ (line \ref{lin.driftdetection} to line \ref{lin.classifiers pool}).
If the received example belongs to the target domain, all  weights of all sub-classifiers $\omega_{h^{j,k}_i}$, $\forall i, j, k$,  are reset (line \ref{lin.drfit.reweight}), to re-identify which sub-classifiers match the current target distribution well. 

After checking for concept drift, the most recent ensemble $H_i^{J_i}$ created for the source or target $i$ is trained on the current example (line \ref{lin.onlineboosting.target domain}).  
If the example belongs to the target (line \ref{lin.target received}), it is used to update the weight $\omega_{h^{j,k}_i}$ of each sub-classifier  (line \ref{lin.assign weight.target.start} to line \ref{lin.assign.weight.target.end}). The weight of each subclassifier is proportional to its accuracy on the target examples, giving exponentially less importance to older examples. How much less importance is controlled based on a pre-defined parameter $\theta$, $0 \leq \theta \leq 1$. The use of $\theta$ helps to deal with non-stationary environments, and with the fact that source ensembles may be updated on new examples before a given target example is received. Weight calculation is explained next.

$A_{h^{j,k}_i}^{(t)}$ represents how well a sub-classifier performs on the target (line \ref{lin.time forgetting factor}).  When $t = 1$, $A_{h^{j,k}_i}^{(1)} = 1 - L(h^{j,k}_i(x^{(1)}))$, where $L(h^{j,k}_i(x^{(1)}))$ calculated based on the probabilities given by the sub-classifiers. 
When the next target examples are received ($t > 1$), we use the time forgetting factor to multiply the previous value of $A_{h^{j,k}_i}^{(t)}$. Therefore, $\theta$ can reduce the contribution of older data and increase the importance of newer data. 

After that, we let $A_{h^{j,k}_i}^{(t)}$ be divided by the normalisation factor $\sum_{t' = 1}^t \theta^{(t'-1)}$ (line \ref{lin.alpha}). Thus, $\alpha_{h^{j,k}_i}^{(t)}$ (line \ref{lin.alpha}) represents the current performance of each sub-classifier through a value between 0 and 1. This enables us to interpret this performance as a percentage, to decide whether to use or not to use a given sub-classifier for predictions to the current target concept. For instance, when dealing with binary classification problems, we will not use any learner whose accuracy is worse than that of a random classifier for making predictions. This is done by assigning weight zero to any sub-classifier associated to $\alpha_{h^{j,k}_i}^{(t)} <= \lambda$, where $\lambda = 0.5$  (line \ref{lin.assign.weight.target.end}). The weights of all sub-classifiers associated to $\alpha_{h^{j,k}_i}^{(t)} > \lambda$ are set to their predictive performance $\alpha_{h^{j,k}_i}^{(t)}$ normalised by the sum of the predictive performances of all the sub-classifiers associated to $\alpha_{h^{j,k}_i}^{(t)} > \lambda$ (line \ref{lin.assign.weight.target.end}). 

This means that any sub-classifier that is incompatible with the current target is prevented from being used for predictions, avoiding negative transfer. The other sub-classifiers all contribute towards predictions, i.e., they are all used to transfer knowledge to the current target. The extent to which they contribute is determined by their weight.

When a prediction needs to be made, we multiply the corresponding weights of the sub-classifiers with the probabilistic prediction made by each sub-classifier. All sub-classifiers $h^{j,k}_i$, $i\in M$, $j \in \{1,\cdots, J_i\}$, $k \in \{1,\cdots,K\}$ are considered for this. Afterwards, we get the sum of the weighted predicted probabilities of each class and use majority vote to decide the predicted class. 


\section{Experiment Setup}
\label{experiment setup}

This paper aims to answer the following research question: can  multi-source  transfer learning  improve  the  predictive performance in data stream mining? When and why? For that, we proposed Melanie. We now present the setup of experiments made to answer this research question through Melanie. 

\subsection{Data Sets}
\label{sec:datasets}


\subsubsection{Artificial Data Sets}

The artificial data sets consist of two real value input variables and a binary output. 
Each class in a given source or target is associated to a Gaussian distribution. Three different target scenarios are generated by varying the number of target training examples of each class (class size) in \{50, 500, 5000\}, simulating small, medium and large sample size. All the source data sets have 5000 examples for each class. We evaluate the algorithm under three different situations (no concept drift, abrupt concept drift, and incremental concept drift). All source training data was used for training before the target data started to be presented. The parameters of the Gaussians used to create the data sets are presented in Section \ref{experiment results}, together with the analysis of the results of the experiments that use them.




\subsubsection{Real World Data Sets}

Two widely used real world data sets Electricity (ELEC2) \cite{gama2004learning,harries1999splice} and Weather  \cite{elwell2011incremental, ditzler2013incremental} were used. ELEC2 has 5 numeric input features and one binary output, and contains 45312 examples. Weather has 8 numeric input features and one binary output, and contains 18159 examples. Both data sets are likely to contain concept drifts, given the conditions under which they were generated. Further details on these data sets are omitted due to space restrictions, and can be found at \cite{gama2004learning,harries1999splice,elwell2011incremental, ditzler2013incremental}.

To simulate if the sources do or do not share the same concept as the target, we extracted some examples from the real world data sets in two ways.
First, to keep the original distribution, we randomly extract $P_S =$ 30\%, 60\%, 90\% of the instances for each class label in each day from the ELEC2 data set as the source domain. For instance, each day has 48 examples. If UP label has 15 examples and DOWN label has 33 examples in one day, we randomly pick 4, 9, 13 instances from UP label and 9, 19, 29 instances from DOWN label to compose three source domain data sets representing three different evaluation scenarios. The target domain has the rest $P_T$ percentage of the data. For the weather data set, we extract $P_S=$ 30\%, 60\%, 90\% of the examples for each class label in each month as the source domain and use the rest $P_T$ percentage of the data as the target. All the instances keep their original chronological order. This way can also simulate the case where both source and target are producing data over time. 
Second, to simulate the case where the source and target do not share the same concept, we extract the first $P_S=$ 30\%, 60\%, 90\% instances of the data sets. The rest $P_T$\% of the data composes the target domain.



\subsection{Compared Approaches and Parameter Choice}

In order to check whether multi-source transfer learning can improve predictive performance in data stream mining, we compared the following approaches: 

\begin{itemize}
\item \textbf{Melanie:} Online Bagging \cite{oza2005online} and Online Boosting \cite{oza2005online} were investigated as Melanie's ensemble learning approaches, and DDM \cite{kolter2007dynamic} was used as the drift detection method. These approaches have been chosen due to their popularity. Other online learning approaches and drift detection methods can be investigated as future work. 

\item \textbf{Melanie without any sources:} this is the same as Melanie, but without using any sources. It will enable us to know whether Melanie is able to benefit from sources. 

\item \textbf{Existing data stream learning approaches for non-stationary environments:} Dynamic Weighted Majority (DWM) \cite{kolter2007dynamic}, Adaptive Random Forest (ARF) \cite{gomes2017adaptive}, DDM \cite{kolter2007dynamic} with Online Bagging, and DDM with Online Boosting were compared against Melanie. This enables us to know to what extent transfer learning can be helpful in view of existing approaches for dealing with concept drift. The first two are widely used approaches, available in the MOA \cite{DBLP:journals/jmlr/BifetHKP10} framework. The latter two make use of the same base ensemble learning algorithms and drift detection method as Melanie, helping us to check whether Melanie's use of multi-sources is beneficial.  

\item \textbf{Baselines:} Online Bagging  and Online Boosting \cite{oza2005online}, which do not have mechanisms to cope with concept drift.  
\end{itemize}

The sub-classifiers of all approaches were Hoeffding Trees \cite{domingos2000mining} except for ARF, which uses a variation of Hoeffding Tree called ARFHoeffding Tree \cite{gomes2017adaptive, DBLP:journals/jmlr/BifetHKP10}. Other sub-classifiers will be investigated as future work. To facilitate the comparisons and create readable plots to compare accuracies over time, the comparisons are separated into three groups: (1) approaches using Online Bagging, (2) approaches using Online Boosting and (3) Melanie against approaches that are not based on Online Bagging or Online Boosting. 


For all the approaches, we chose parameters based on grid search. For Melanie, we investigated $\theta = 0:0.1:1$ and $\delta$ = 0.05. The $\lambda$ value is set to 0.5, as we are dealing with binary classification problems. 
For Online Bagging and Online Boosting, the size of the sub-classifiers is varied in 1:1:30. For DWM, $\beta$ was investigated in 0:0.1:1, period $p$ = 1, and weight threshold for removing sub-classifiers 0.01. For ARF, the number of trees is in 10:1:30 (MOA restricts minimum ARF ensemble size as 10).   

\subsection{Performance Metrics}

The performance of the compared approaches was measured based on the accuracy on the target examples. 
When using artificial data sets, the accuracy was calculated in a prequential way and was reset to zero upon concept drift \cite{minku2012ddd}. This enables us to measure the performance on each concept separately, without being affected by the performance on the previous concepts. 
For the real world data sets, as we do not know when concept drift happens, accuracy was calculated over a sliding window \cite{gama2009issues} whose size is a percentage of the data stream, corresponding to the percentage used in \cite{kolter2007dynamic}. 

All stochastic approaches (which are all approaches except for DWM) were run 30 times, and the average accuracy across these 30 runs is reported.

Friedman tests on each data set were used to check if there is significant difference between any pair of approaches. If there is, Nemenyi Post-Hoc test was used to identify which pair of approaches is really different from each other.

\section{Experiment Results}
\label{experiment results}

This section presents the results of the experiments on artificial (Section \ref{results:artificial}) and real world (Section \ref{results:real}) data sets. Table \ref{tab.rs.rank} presents the rank of each approach on each data set.

\subsection{Experiments on Artificial Data}
\label{results:artificial}

\begin{table}[t]
\caption{Multi-Source Data Set Distributions.}
\begin{center}
\scalebox{0.75}{
\begin{tabular}{|c|c|c|c|}
\hline
\textbf{Domain}&\textbf{Class} & \textbf{\textit{Center}}& \textbf{\textit{Covariance matrix}} \\
\hline
Target&Class 0& $(2,3)$&$\begin{pmatrix}
2 & 0\\
0 & 2
\end{pmatrix}$ \\
\cline{2-4}
&Class 1& $(7,8)$&$\begin{pmatrix}
2 & 0\\
0 & 2
\end{pmatrix}$\\
\hline
Source 1&Class 0& $(-3,6)$&$\begin{pmatrix}
3 & 0\\
0 & 2
\end{pmatrix}$ \\
\cline{2-4}
&Class 1& $(7,8)$&$\begin{pmatrix}
3 & 0\\
0 & 2
\end{pmatrix}$\\
\hline
Source 2&Class 0& $(2,1)$&$\begin{pmatrix}
1 & 0\\
0 & 2
\end{pmatrix}$ \\
\cline{2-4}
&Class 1& $(7,8)$&$\begin{pmatrix}
1 & 0\\
0 & 2
\end{pmatrix}$\\
\hline
\end{tabular}}
\label{tab.multi-sources data sets parameters}
\end{center}
\end{table}

\begin{figure}[t]
\begin{subfigure}{0.24\textwidth}
\centerline{\includegraphics[scale=0.21]{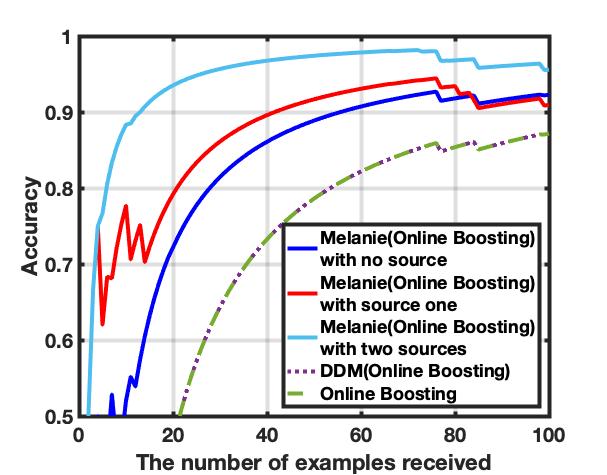}}
\caption{Each class size is 50}
\label{fig.e150boost}
\end{subfigure}
\begin{subfigure}{0.24\textwidth}
\centerline{\includegraphics[scale=0.21]{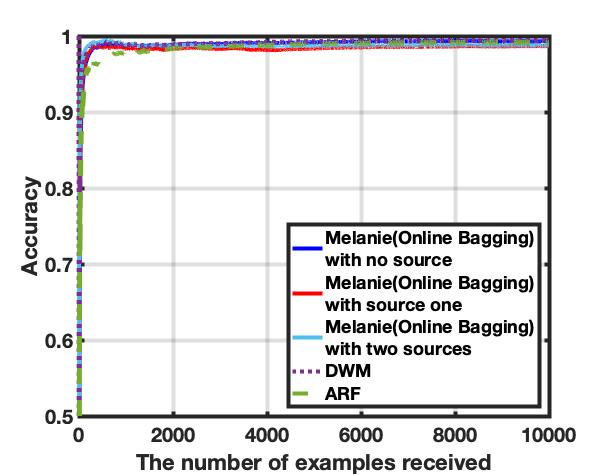}}
\caption{Each class size is 5000}
\label{fig.e15000others}
\end{subfigure}
\caption{Accuracy on data sets with no concept drift.}
\label{fig.e1}

\end{figure}

\begin{table*}[t]
\begin{threeparttable}[t]
\caption{Friedman Ranks on Each Data Set.}
\label{tab.E1.rank}
\label{tab.E2s.rank}
\label{tab.E2i.rank}
\label{tab.rs.rank}
\label{tab.ws.rank}
\renewcommand{\arraystretch}{0.9}
\renewcommand{\tabcolsep}{2.5pt}
\label{tab:my_label}
\begin{center}
\begin{tabular}{|l|c|c|c|c|c|c|c|c|c|c|c|c|c|c|c|c|c|c|c|c|c|}  
\hline
\multirow{2}{*}{Data Set}
&\multicolumn{3}{c|}{\multirow{2}{*}{No Drift}}
&\multicolumn{3}{c|}{\multirow{2}{*}{Abrupt}}
&\multicolumn{3}{c|}{\multirow{2}{*}{Incremental}}
&\multicolumn{6}{c|}{ELEC2}
&\multicolumn{6}{c|}{Weather}\\\cline{11-22}

&\multicolumn{3}{c|}{}
&\multicolumn{3}{c|}{}
&\multicolumn{3}{c|}{}
&\multicolumn{3}{c|}{Similar}
&\multicolumn{3}{c|}{Non-similar}
&\multicolumn{3}{c|}{Similar}
&\multicolumn{3}{c|}{Non-similar}
\\
\hline
Class size or $P_S$
&50&500&5000
&50&500&5000
&50&500&5000
&0.9&0.6&0.3
&0.9&0.6&0.3
&0.9&0.6&0.3
&0.9&0.6&0.3
\\
\hline
\textbf{Melanie(Online Bagging) without source}&7.3&4.5&2.3&7.8&5.1&4.1&5.7&5.4&4.8&7.4&5.2&3.6&3.9&2.8&3.6&5.6&7.0&7.5&6.9&5.9&6.0\\
\hline
\textbf{Melanie(Online Bagging) with source one}&3.8&\textbf{1.7}&9.0&-&-&-&-&-&-&-&-&-&-&-&-&-&-&-&-&-&-\\
\hline
\textbf{Melanie(Online Bagging) with all sources}&\textbf{2.0}&3.5&6.2&\textbf{2.1}&\textbf{1.0}&\textbf{1.3}&\textbf{2.7}&\textbf{2.0}&\textbf{2.8}&4.5&2.2&2.7&\textbf{2.3}&\textbf{2.5}&\textbf{2.2}&7.4&6.5&8.2&4.4&5.3&5.2\\
\hline
\textbf{Melanie(Online Boosting) without source}&5.0&8.3&6.1&3.1&3.3&5.3&5.9&7.1&7.2&4.4&\textbf{1.6}&2.1&2.6&\textbf{2.4}&\textbf{2.2}&5.5&5.8&5.3&5.2&6.3&8.2\\
\hline
\textbf{Melanie(Online Boosting) with source one}&3.6&7.9&11.8&-&-&-&-&-&-&-&-&-&-&-&-&-&-&-&-&-&-\\
\hline
\textbf{Melanie(Online Boosting) with all sources}&\textbf{1.1}&5.4&10.4&\textbf{1.2}&2.0&3.1&\textbf{3.1}&3.2&3.9&8.2&2.4&\textbf{1.9}&\textbf{2.2}&\textbf{2.4}&\textbf{2.1}&4.3&2.3&4.4&\textbf{2.8}&4.0&8.0\\
\hline
\textbf{DDM(Online Bagging)}&8.2&6.0&3.7&8.6&7.1&5.9&5.5&5.7&4.5&6.4&9.6&9.3&9.3&9.3&9.4&6.0&7.2&6.8&7.4&8.3&6.7\\
\hline
\textbf{DDM(Online Boosting)}&11.4&10.9&8.8&4.9&8.8&8.8&6.8&6.5&6.6&4.3&6.9&7.2&7.3&7.1&7.1&\textbf{3.3}&4.7&3.9&5.9&5.1&3.4\\
\hline
\textbf{Online Bagging}&8.2&6.0&3.7&9.0&9.9&10.0&8.0&7.0&6.8&6.6&9.0&9.6&9.2&9.7&9.6&6.0&6.7&6.0&7.0&6.5&6.0\\
\hline
\textbf{Online Boosting}&11.4&11.8&8.8&4.9&5.9&7.0&8.2&8.8&9.0&4.3&7.6&7.7&7.1&7.8&7.8&\textbf{3.3}&4.1&3.0&5.4&4.4&4.0\\
\hline
\textbf{Dynamic Weighted Majority}&9.5&2.8&\textbf{1.2}&6.8&4.8&1.9&4.3&3.5&3.5&6.6&5.7&5.4&6.0&5.8&5.6&9.7&9.1&8.4&5.9&7.5&6.5\\
\hline
\textbf{Adaptive Random Forest}&6.3&9.3&6.0&6.7&7.0&7.5&5.0&5.8&6.0&\textbf{2.3}&4.6&5.4&5.0&5.2&5.5&3.9&\textbf{1.7}&\textbf{1.6}&4.1&\textbf{1.6}&\textbf{1.1}\\
\hline

\end{tabular}
\begin{tablenotes}
\item Friedman's p-values were always $<2.2 \times 10^{-16}$. The best approach and the approaches not significantly different from it according to the Nemenyi test are in bold.
\end{tablenotes}
\end{center}
\end{threeparttable}
\vspace{-0.45cm}
\end{table*}

\subsubsection{Multi-sources effect}
\label{multi-source.effect}

This experiment aims to investigate whether the use of different sources by Melanie can help to improve accuracy under different amounts of target training data, when dealing with stationary environments. In particular, we would like to test the hypothesis that Melanie can benefit from multiple sources to improve accuracy when there is a lack of source training data. We would also like to check whether or not they the use of sources could be detrimental to accuracy when there is abundant target training data, or when sources do not match the target exactly. 
Table \ref{tab.multi-sources data sets parameters} lists the distributions of the target and source domains used in this experiment. 

The Friedman ranking of the approaches on each no drift data set is shown in Table \ref{tab.E1.rank}. Figure \ref{fig.e1} shows two representative results across time. Other figures were omitted due to space restrictions. When each class size was 50, Melanie with two sources obtained the best performance, followed by Melanie with one source and no source.
The fact that the sources did not match the target exactly was not detrimental to Melanie's accuracy.

The more target examples were received, the more similar the accuracy of the approaches became (see e.g., Figure \ref{fig.e150boost}). When the class size was 500, Melanie with sources still obtained competitive ranking (see Table \ref{tab.E1.rank}). When the class size was 5000, Melanie obtained worse ranking than other approaches such as DWM. However, the magnitude of the differences in accuracy among all approaches was very small (see e.g., Figure \ref{fig.e15000others}). Therefore, even though Melanie had a detrimental effect, this detrimental effect was very small.

These experiments show that Melanie was able to benefit from different sources, and this was particularly helpful during the periods where there is not enough target data to learn from. Once the amount of target data becomes sufficient, source data becomes unnecessary.

It is also worth noting that, since the data in this experiment have no concept drift, Melanie without source and DDM usually had the same sub-classifiers as Online Bagging or Online Boosting. 
And yet, Melanie (Online Boosting) without source still outperformed DDM (Online Boosting) and Online Boosting, for all class sizes. The differences in accuracy were statistically significant according to Nemenyi tests. The same is valid when using Online Bagging for class sizes of 500 and 5000. As the main difference between Melanie without sources and these other approaches is its weighting strategy, this suggests that Melanie's weighting strategy is more adequate.

\subsubsection{Abrupt Concept Drift}

\begin{figure}[t]
\begin{subfigure}{0.24\textwidth}
\centerline{\includegraphics[scale=0.21]{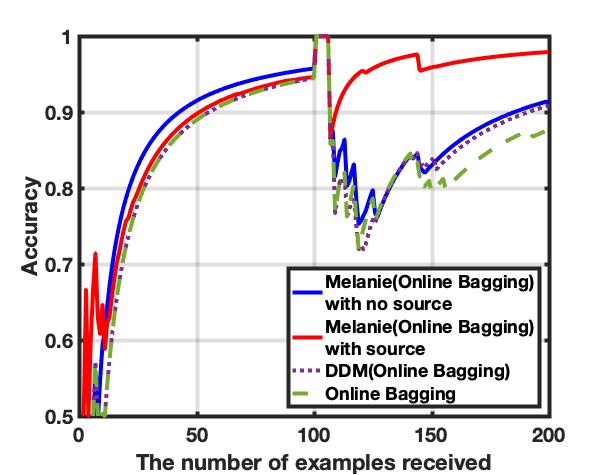}}
\caption{Abrupt; class size of 50}
\label{fig.e2s50bag}
\end{subfigure}
\begin{subfigure}{0.24\textwidth}
\centerline{\includegraphics[scale=0.21]{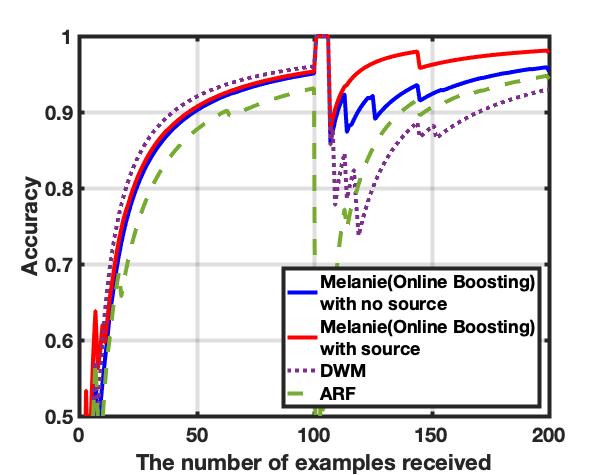}}
\caption{Abrupt; class size of 50}
\label{fig.e2s50other}
\end{subfigure}

\begin{subfigure}{0.24\textwidth}
\centerline{\includegraphics[scale=0.21]{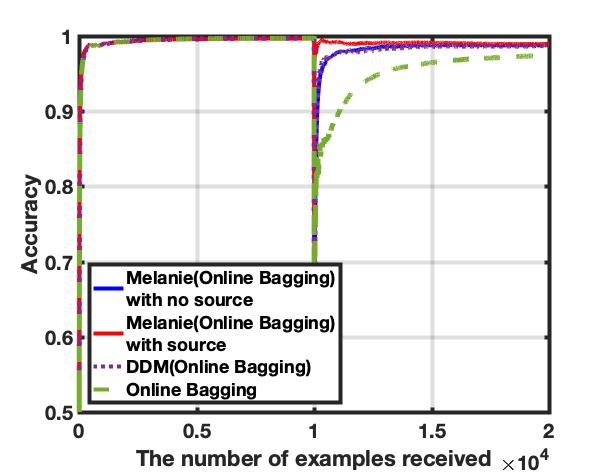}}
\caption{Abrupt; class size of 5000}
\label{fig.e2s5000bag}
\end{subfigure}
\begin{subfigure}{0.24\textwidth}
\centerline{\includegraphics[scale=0.21]{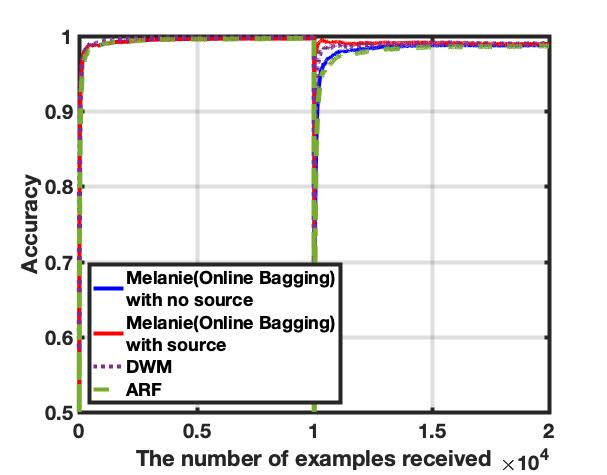}}
\caption{Abrupt; class size of 5000}
\label{fig.e2s5000others}
\end{subfigure}

\begin{subfigure}{0.24\textwidth}
\centerline{\includegraphics[scale=0.21]{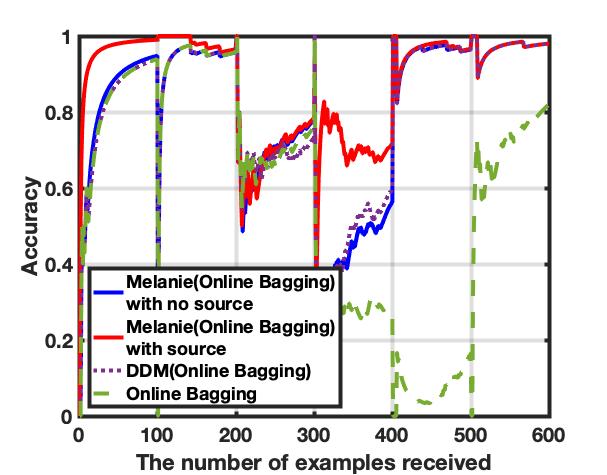}}
\caption{Incremental; class size of 50}
\label{fig.e2i50bag}
\end{subfigure}
\begin{subfigure}{0.24\textwidth}
\centerline{\includegraphics[scale=0.21]{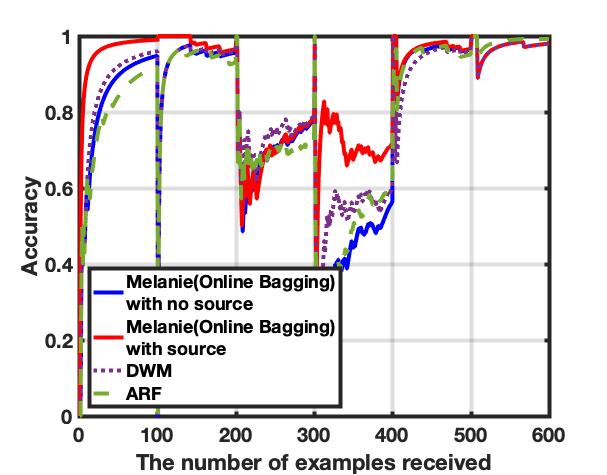}}
\caption{Incremental; class size of 50}
\label{fig.e2i50others}
\end{subfigure}

\begin{subfigure}{0.24\textwidth}
\centerline{\includegraphics[scale=0.21]{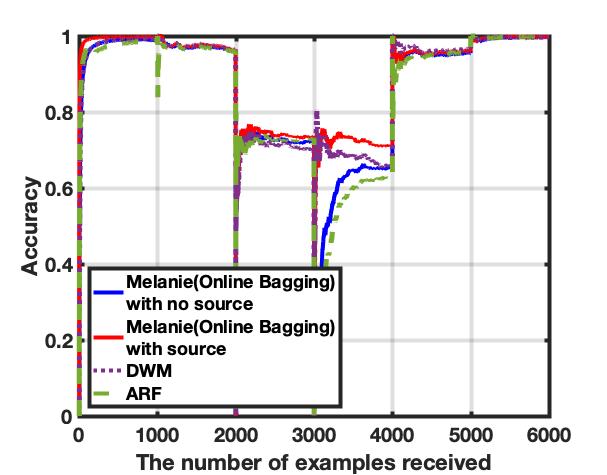}}
\caption{Incremental; class size of 500}
\label{fig.e2i500others}
\end{subfigure}
\begin{subfigure}{0.24\textwidth}
\centerline{\includegraphics[scale=0.21]{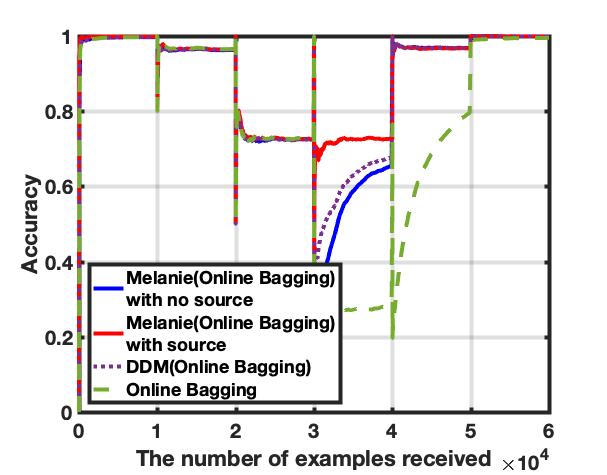}}
\caption{Incremental; class size of 5000}
\label{fig.e2i5000bag}
\end{subfigure}

\caption{Accuracy on abrupt and incremental concept drift data.}
\label{fig.e2s}
\label{fig.e2i}
\vspace{-0.5cm}
\end{figure}

This experiment considers that the target data streams have one abrupt concept drift in the middle of the target data stream, and the source concept follows the distribution of the target concept that is valid after the drift. It enables us to check whether Melanie with this source is able to obtain good accuracy by identifying that this source is useful after the drift, and by preventing any detrimental effect that could potentially be caused by using it before the drift, when it does not match the target well. Table \ref{tab.abrupt drift data sets parameters} shows the parameters of the abrupt drift data sets.

Based on Friedman and Nemenyi tests (Table \ref{tab.E2s.rank}), we can see that Melanie with source presented the best results over all the abrupt drift data sets. 
Larger improvements in accuracy occurred mainly in the beginning of the learning period and after the drift (see e.g., Figures \ref{fig.e2s50bag}, \ref{fig.e2s50other}, \ref{fig.e2s5000bag} and \ref{fig.e2s5000others}). 
Similar to Section \ref{multi-source.effect}, the more target examples were received, the more similar the accuracies of different approaches became, meaning that the use of different sources is helpful during the periods when target examples are not abundant. This is an encouraging result, which demonstrates that Melanie can speed up recovery from concept drift. In particular, it managed to speed up recovery from concept drift in comparison to other approaches specifically designed for non-stationary environments, such as DDM, DWM and ARF.

Sometimes, DWM obtained slightly better accuracy than Melanie with source before the drift, after enough examples from the target concept were received (see e.g., Figure \ref{fig.e2s50other}). However, the improvement in accuracy was very small compared to the benefit provided by Melanie with source in the beginning of the learning period and after concept drifts.

Overall, Melanie with source was particularly helpful to speed up adaptation to new concepts. 

\begin{table}[t]
\caption{Abrupt Concept Drift Data Sets Distributions.}
\vspace{-0.3cm}
\renewcommand{\arraystretch}{0.9}
\begin{center}
\scalebox{0.75}{
\begin{tabular}{|c|c|c|c|}
\hline
\textbf{Domain}&\textbf{Class} & \textbf{\textit{Center}}& \textbf{\textit{Covariance matrix}} \\
\hline
Target before Concept Drift&Class 0& $(2,3)$&$\begin{pmatrix}
1 & 0\\
0 & 2
\end{pmatrix}$ \\
\cline{2-4}
&Class 1& $(7,8)$&$\begin{pmatrix}
1 & 0\\
0 & 2
\end{pmatrix}$\\
\hline
Target after Concept Drift&Class 0& $(2,9)$&$\begin{pmatrix}
1 & 0\\
0 & 2
\end{pmatrix}$ \\
\cline{2-4}
&Class 1& $(5,4)$&$\begin{pmatrix}
1 & 0\\
0 & 2
\end{pmatrix}$\\
\hline
Source &Class 0& $(2,9)$&$\begin{pmatrix}
1 & 0\\
0 & 2
\end{pmatrix}$ \\
\cline{2-4}
&Class 1& $(5,4)$&$\begin{pmatrix}
1 & 0\\
0 & 2
\end{pmatrix}$\\
\hline
\end{tabular}}
\label{tab.abrupt drift data sets parameters}
\end{center}
\vspace{-0.5cm}
\end{table}


\subsubsection{Incremental Concept Drift}

The parameters of the incremental concept drift data sets are shown in Table \ref{tab.incremaental drift data sets parameters}. For the class sizes of 50, 500, 5000, at each 100, 1000, and 10000 time steps, the centres of the Gaussian of class 0 and class 1 move towards each other by 1 unit, until the Gaussians of class 0 and 1 swap location. Six different sources are available, one corresponding to each intermediate concept between the original concept and the new concept. The aim is to check whether Melanie can identify which source models to emphasise, to improve predictive performance during and right after the drift. 

Based on Friedman and Nemenyi (Table \ref{tab.E2i.rank}), we can see that Melanie (Online Bagging) with source performs best on incremental drift data sets after concept drift, for all target class sizes. Figures \ref{fig.e2i50bag}, \ref{fig.e2i50others}, \ref{fig.e2i500others} and \ref{fig.e2i5000bag} show representative examples of Melanie (Online Bagging) with source's outperforming accuracy. 
This shows that Melanie can be frequently helpful to recover from gradual drifts, given that not enough examples belonging to intermediate target concepts will be received for approaches to learn them well without knowledge transfer.


\begin{table}[t]
\caption{Incremental Concept Drift Data Sets Distributions.}
\vspace{-0.3cm}
\begin{center}
\scalebox{0.75}{
\begin{tabular}{|c|c|c|c|}
\hline
\textbf{Domain}&\textbf{Class} & \textbf{\textit{Center}}& \textbf{\textit{Covariance matrix}} \\
\hline
Target before concept&Class 0& $(2,3)$&$\begin{pmatrix}
1 & 0\\
0 & 2
\end{pmatrix}$ \\
\cline{2-4}
&Class 1& $(7,8)$&$\begin{pmatrix}
1 & 0\\
0 & 2
\end{pmatrix}$\\
\hline
Target after concept drift&Class 0& $(7,8)$&$\begin{pmatrix}
1 & 0\\
0 & 2
\end{pmatrix}$ \\
\cline{2-4}
&Class 1& $(2,3)$&$\begin{pmatrix}
1 & 0\\
0 & 2
\end{pmatrix}$\\
\hline

Source 1&Class 0& $(2,3)$&$\begin{pmatrix}
1 & 0\\
0 & 2
\end{pmatrix}$ \\
\cline{2-4}
&Class 1& $(7,8)$&$\begin{pmatrix}
1 & 0\\
0 & 2
\end{pmatrix}$\\
\hline
Source 2&Class 0& $(3,4)$&$\begin{pmatrix}
1 & 0\\
0 & 2
\end{pmatrix}$ \\
\cline{2-4}
&Class 1& $(6,7)$&$\begin{pmatrix}
1 & 0\\
0 & 2
\end{pmatrix}$\\
\hline
Source 3&Class 0& $(4,5)$&$\begin{pmatrix}
1 & 0\\
0 & 2
\end{pmatrix}$ \\
\cline{2-4}
&Class 1& $(5,6)$&$\begin{pmatrix}
1 & 0\\
0 & 2
\end{pmatrix}$\\
\hline
Source 4&Class 0& $(5,6)$&$\begin{pmatrix}
1 & 0\\
0 & 2
\end{pmatrix}$ \\
\cline{2-4}
&Class 1& $(4,5)$&$\begin{pmatrix}
1 & 0\\
0 & 2
\end{pmatrix}$\\
\hline
Source 5&Class 0& $(6,7)$&$\begin{pmatrix}
1 & 0\\
0 & 2
\end{pmatrix}$ \\
\cline{2-4}
&Class 1& $(3,4)$&$\begin{pmatrix}
1 & 0\\
0 & 2
\end{pmatrix}$\\
\hline
Source 6&Class 0& $(7,8)$&$\begin{pmatrix}
1 & 0\\
0 & 2
\end{pmatrix}$ \\
\cline{2-4}
&Class 1& $(2,3)$&$\begin{pmatrix}
1 & 0\\
0 & 2
\end{pmatrix}$\\
\hline
\end{tabular}}
\label{tab.incremaental drift data sets parameters}
\end{center}
\vspace{-0.5cm}
\end{table}



\subsection{Experiments on Real-World Data}
\label{results:real}

\subsubsection{ELEC2 Data}

\begin{figure}[t]
\begin{subfigure}{0.24\textwidth}
\centerline{\includegraphics[scale=0.21]{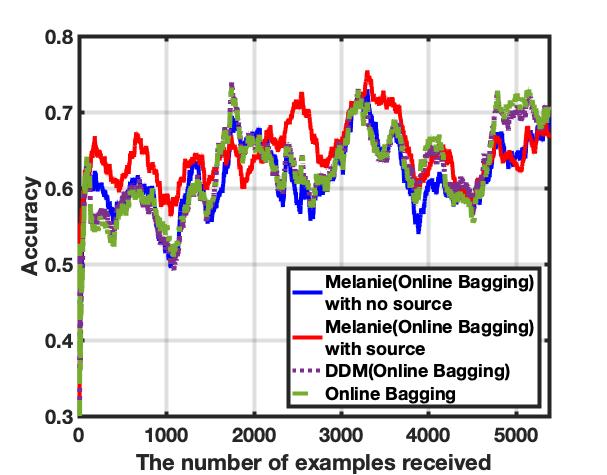}}
\caption{ELEC2 similar; $P_S = 0.9$}
\label{fig.rs9bag}
\end{subfigure}
\begin{subfigure}{0.24\textwidth}
\centerline{\includegraphics[scale=0.21]{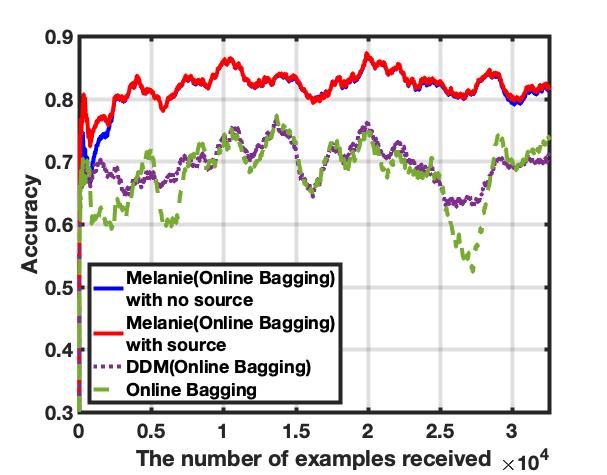}}
\caption{ELEC2 similar; $P_S=0.3$}
\label{fig.rs3bag}
\end{subfigure}
\begin{subfigure}{0.24\textwidth}
\centerline{\includegraphics[scale=0.21]{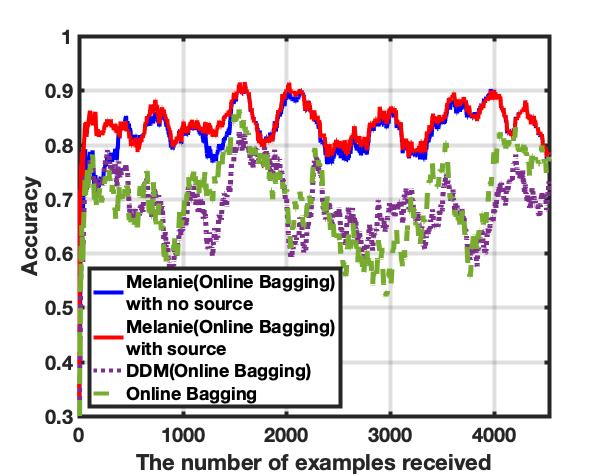}}
\caption{ELEC2 non-similar; $P_S = 0.9$}
\label{fig.rns9bag}
\end{subfigure}
\begin{subfigure}{0.24\textwidth}
\centerline{\includegraphics[scale=0.21]{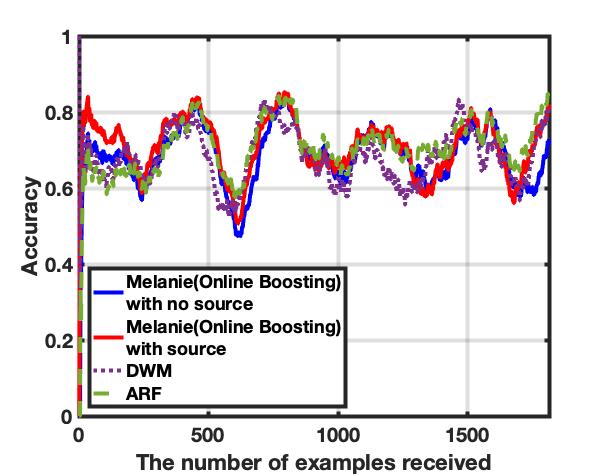}}
\caption{Weather non-similar; $P_S$=0.9}
\label{fig.wns9others}
\end{subfigure}
\caption{Accuracy on ELEC2 and Weather data.}
\label{fig.rs}
\vspace{-0.5cm}
\end{figure}


Based on Friedman and Nemenyi tests (Table \ref{tab.rs.rank}), Melanie (Online Boosting) with source and Melanie (Online Bagging) with source hold the best and second best performance on over all ELEC2 data sets with non-similar source. Melanie without source was also competitive. 
For data sets with similar source, Melanie (Online Boosting) with and without source achieved the top two accuracies when $P_S$ was 0.3 and 0.6. Figures  \ref{fig.rns9bag}, \ref{fig.rs9bag} and \ref{fig.rs3bag} show some representative results. 

The probable reason for the good results achieved by Melanie is that concept drifts are likely to occur very frequently in this data set \cite{minku2012ddd}, causing the number of target examples from a given concept to be relatively small even for the cases with smaller $P_S$. 
As with Section \ref{multi-source.effect}, using dissimilar sources still helped to improve accuracy. 
Moreover, the fact that Melanie without source was competitive on a data set likely to contain concept drifts also indicates that Melanie's maintenance of old target sub-classifiers can be helpful to deal with drifts.
These results also demonstrate that Melanie can not only enable learning over time in the target, but also in the source domain. 

\subsection{Weather Data}



Based on Friedman and Nemenyi tests (Table \ref{tab.ws.rank}), ARF performed best when $P_T$ was larger. 
Overall, Melanie was not helpful but not really much detrimental either, as the magnitude of the differences in accuracy between Melanie and the top ranked approaches were small. An example of representative result is shown in Figure \ref{fig.wns9others}.
Still, Melanie with source preformed better in beginning of the learning period in most cases with similar sources and in all cases with non-similar sources. The probable reason for Melanie not to outperform others is the fact that no drift detections were performed by DDM in this data set. The high variance of accuracy throughout the learning could mean that no concept drifts that are more significant than the inherent variability and noise in the training examples are present. Therefore, source data is only useful in the beginning of the learning period, when there are not enough target training examples.

\begin{table}[t]
\begin{center}
\caption{Summary of Friedman rank with drift data sets.}
\scalebox{1}{
\begin{tabular}{|l|c|}
\hline
\textbf{Approaches}&\textbf{Average Rank}\\
\hline
\textbf{Melanie(Online Bagging) without source}&
5.46\\
\hline
\textbf{Melanie(Online Bagging) with source}&
3.63\\
\hline
\textbf{Melanie(Online Boosting) without source}&
4.64\\
\hline
\textbf{Melanie(Online Boosting) with source}&
\textbf{3.42}\\
\hline
\textbf{DDM(Online Bagging)}&
7.39\\
\hline
\textbf{DDM(Online Boosting)}&
6.03\\
\hline
\textbf{Online Bagging}&
7.92\\
\hline
\textbf{Online Boosting}&
6.13\\
\hline
\textbf{Dynamic Weighted Majority}&
5.94\\
\hline
\textbf{Adaptive Random Forest}&
4.45\\
\hline
\end{tabular}}
\label{tab.summary.rank}
\end{center}
\vspace{-0.5cm}
\end{table}

\subsection{Summary and Answer to the Research Question}


Table \ref{tab.summary.rank} is the summary of Friedman rank of each approach in each data set. We can see that Melanie (Online Boosting) with source has the best average rank across data sets. 
%
Overall, our experiments show that multi-sources can be helpful for improving accuracy in data stream mining. The more similar the source and target domains are, and the smaller the number of target training examples, the larger the benefit provided by multi-sources. The use of multi-sources was able to speed up recovery from concept drift, by leading to better accuracy especially during and right after the drifts. This is because Melanie was able to identify and benefit from cases when the source sub-concepts matched the target, enabling predictions to be performed based on extra data that are compatible with the target or with intermediate concepts. Such data is only necessary when there are not enough target training examples representing the current target or intermediate concept.





\section{Conclusion and Future Work}
\label{conclusion}
This work introduced multi-source transfer learning for non-stationary environments, and proposed the first approach (Melanie) able to transfer knowledge between different data streaming sources and a data streaming target.


We performed experiments with several different data sets to evaluate Melanie and check whether multi-sources can be beneficial to improve accuracy in data stream mining. The results show Melanie  can transfer and pick the most suitable knowledge under variety of scenarios with and without concept drift, improving accuracy especially during periods of time when there is not a large amount of target training data. As such, multi-source transfer learning can help to speed up adaptation to concept drift. Melanie was also usually able to avoid negative transfer.

Future work includes a sensitivity analysis of Melanie's parameters; an investigation of Melanie with other types of sub-classifiers, online learning ensembles, drift detection methods and data sets; and an extension of Melanie to tackle class imbalance.

\section*{Acknowledgements}

This work was supported by EPSRC Grant Nos.~EP/R006660/1 and EP/R006660/2.

\bibliographystyle{IEEEtran}
\bibliography{References}

\end{document}